\documentclass[10pt]{article}
\usepackage{amsmath}    
\usepackage{amsfonts}
\usepackage{amssymb}  
\usepackage{bigints}
\usepackage{wrapfig}
\usepackage{graphicx}   
\usepackage{subfig}
\usepackage{verbatim}   
\usepackage{color}      
\usepackage{parskip}
\usepackage{float}
\usepackage{courier}
\usepackage{exercise}
\usepackage{sistyle}
\SIthousandsep{,}
\usepackage{makeidx}
\makeindex
\usepackage[nottoc]{tocbibind}

\usepackage[colorlinks = true,
          linktocpage=true,
            pagebackref=true, 
            linkcolor = red,
            urlcolor  = blue,
            citecolor = red,
            anchorcolor = blue]{hyperref}
\definecolor{dkgreen}{rgb}{0,0.6,0}
\definecolor{gray}{rgb}{0.5,0.5,0.5}
\definecolor{mauve}{rgb}{0.58,0,0.82}
\definecolor{index}{rgb}{0.88,0.32,0}

\usepackage{listings}
\usepackage{blindtext}
\usepackage{geometry}
\renewcommand{\arraystretch}{1.4} 
\newtheorem{theorem}{Theorem}[section]

\newcommand{\qed}{\nobreak \ifvmode \relax \else
      \ifdim\lastskip<1.5em \hskip-\lastskip
      \hskip1.5em plus0em minus0.5em \fi \nobreak
      \vrule height0.75em width0.5em depth0.25em\fi}

\begin{document}

\hypersetup{linkcolor=blue}

\begin{center}
{\Large \bf{LLMs Without Deep Neural Networks\vspace{0.5ex}\\New Architecture, Benefits \& Case Study} 
}  \\
\addvspace{5ex}
\end{center}
 
\begin{center}
Vincent Granville, Ph.D. $|$  CAIO $|$ vincent@BondingAI.io\\
 \href{https://bondingAI.io/}{BondingAI.io}, version 1.0, May 2026  \\ 
\addvspace{5ex}
\end{center}

\hypersetup{linkcolor=red}



The purpose of this article is to provide validation to my deep neural network alternative
in the context of LLMs. Very recently, there has been a significant interest by Chinese researchers in a model called
RBF network, as a substitute to standard DNNs, with increased explainability
 and higher accuracy. It turns out that my new model, discovered independently, is
 based on the exact same machinery. But with a major twist: it does not need DNN as it finds the
global optimum of the loss function in closed form, in one iteration, thus eliminating the tedious training step.
Here I provide a high-level overview of my technology, with case study and comparison to similar methods.

\section{Building an LLM with alternatives to deep neural networks}

Several approaches have been tried to bypass hard-to-train deep neural networks and replace Blackbox parameters 
with \textcolor{index}{explainable AI} and replicability. Radial basis function (RBF) networks is the most recent one 
to be tested in LLM contexts. It is also the one I pioneered. Some of these methods such as RBF
rely on explainable DNNs. My version of RBF is the only one not using any DNNs, as far as I know. 
 \vspace{1ex}
\begin{itemize}
\item  {\bf Statistical $n$--gram and Index-Based Generation}

Instead of using billions of floating-point numbers to predict text probability, you can model language directly by analyzing a large corpus's frequency and context.
\begin{itemize}
\item  How it works: You build an exact index of every token and its preceding context in your training data. By tallying exactly how often a word follows a specific sequence, the system computes the statistical likelihood of the next token.

\item Tools \& Concepts: This is the foundational concept behind information-retrieval (IR) systems, Markov chains, and probabilistic suffix trees.
\end{itemize}
\vspace{1ex}
\item{\bf Kolmogorov-Arnold Networks (KANs)}

KANs serve as a recent mathematical alternative to Multi-Layer Perceptrons.
\begin{itemize}
\item How it works: While traditional DNNs keep synapse weights fixed and adjust them numerically during training, KANs place learnable functions on the network's edges rather than the nodes. This allows the model to represent highly complex, multi-variable mathematical relationships with smaller architectures.

\item Implementation: You can define KAN structures in Python using the native GitHub - KindXiaoming/pykan library.
\end{itemize}
\vspace{1ex}
\item{\bf $k$-Nearest Neighbors ($k$-NN) / Exact Match Retrievers}

Rather than encoding worldly knowledge into the model's internal matrix weights, you can build a system that acts via dynamic lookup.
\begin{itemize}
\item How it works: During inference, the system searches your pre-indexed training dataset for the chunks of text that most closely match the current context, borrowing from them directly to predict the next word.

\item Tools \& Concepts: To mimic this architecture, you can pair a lightweight parser with high-speed vector databases like Milvus or Qdrant, or use an orchestrator like LangChain for retrieval.
\end{itemize}
\vspace{1ex}
\item{\bf Radial Basis Function (RBF) Networks}

This is a new approach pioneered very recently by several Chinese researchers, and independently by me. My
implementation is the only one not using deep neural networks, allowing for very fast training without epochs or
gradient descent. I discuss it in the remaining of this paper. For details, see chapter 6 in~\cite{vgnewai26}. 
\end{itemize}
\vspace{1ex}

\noindent Both KAN and RBF networks leverage the 
\textcolor{index}{Universal Approximation Theorem} for neural networks. That is, the predictor can approximate any continuous function in any dimension. For KAN, it is a consequence of~the Kolmogorov-Arnold representation theorem. For RBF, it is linked to the universal approximation theorem for Gaussian mixtures.
The basis in RBF models can be a multivatiate Gaussian function leading to a 
\textcolor{index}{Gaussian mixture model}. Even a radial one: that is, spherical rather than ellipsoidal. Thus the name RBF.   
The basis is called kernel in other contexts, and the word \textcolor{index}{kernel method} and RBF can be used interchangeably. 

It is argued that KAN can overfit. In one experiment, KANs fitted the pure random data in the features~to the labels provided, with an extremely high accuracy, see \href{https://medium.com/@rubenszimbres/kolmogorov-arnold-networks-a-critique-2b37fea2112e}{here}. 
This is also true with my implementation of RBF networks. However, in my case, the model benefits from 
\textcolor{index}{benign overfitting} and performs very well outside the training set, even when corrupting the input data with
significant noise. In other words, it acts as a deblurring or high-pass filter even on fairly chaotic data.  

For a recent reference on KAN, see~\cite{kan25a}. For RBF networks, see~\cite{rbf001} discussing GenLoRa and~\cite{rbf003gauss} focusing~on Gaussian basis.  See also~\cite{rbf002sec} focusing on LLM security with radial basis.  
Finally, my recent book about modern AI and LLMs~\cite{vgnewai26} features explainable DNN models as well as RBF networks without DNN.
For a non-technical discussion on the topic, see ``Building LLMs without neural networks",~\href{https://zoea.co.uk/news/news-250127.html}{here}.



\subsection{Connection between RBF networks and standard LLMs}

While Radial Basis Function (RBF) networks and Large Language Models (LLMs) belong to different architectural families, their concepts are converging to improve model efficiency. Recent AI research introduces frameworks like \textcolor{index}{GenLoRA}, which utilize lightweight RBFs to replace large, explicitly stored basis vectors in standard low-rank adaptation methods, achieving superior accuracy on smaller parameter budgets.

Unlike LLM transformers—which build hierarchical, attention-based representations through stacked layers—RBF networks are traditional shallow feedforward networks. However, their ability to perform efficient non-linear function approximation makes them highly relevant when applied alongside LLMs:
\vspace{1ex}
\begin{itemize}
\item {\bf Generative Low-Rank Adapters (GenLoRA):} Instead of explicitly storing bulky basis vectors in matrices like standard LoRA, methods such as GenLoRA use a set of RBF-based non-linear generators to synthesize the necessary basis vectors from a small, shared latent space. This dramatically improves parameter efficiency during LLM fine-tuning.
\vspace{1ex}
\item {\bf Concept Representation \& Interpretability:} RBF Networks are being embedded in low-dimensional spaces for concept visualization. Because they represent non-linear decision boundaries efficiently, they serve as powerful probes to decode and control the internal representations of black-box LLMs.
\vspace{1ex}
\item {\bf Feature Extraction \& Embedding:} The hidden layers of RBF networks compute the Euclidean distance of an input from pre-determined center vectors, often using a Gaussian kernel. While not used as the core architecture of an LLM, RBF components can sit on top of LLM embeddings (e.g., as part of a classifier head) to map dense, high-dimensional textual features to specific, non-linear classification outputs.
\end{itemize}

\subsection{Combining RBF networks with standard LLMs}

Table~\ref{table:kbt54xpq} shows the benefits of combining standard LLMs based on transformers, with the RBF system. 

\begin{table}[H]
\centering
\renewcommand{\arraystretch}{1.4}
\begin{tabular}{|p{3cm}|p{4.5cm}|p{4.8cm}|}
 \hline
   {\bf Feature / Benefit} & {\bf Standard LLM\newline  Transformer} & {\bf RBF Network}\\ [0.5ex] 
 \hline 
\hline 
{\bf Primary Strength} & Capturing deep, sequential\newline context through attention & Highly parameter-efficient\newline non-linear function\newline  approximation \\
\hline
{\bf Computational Overhead} &
High (memory and processing for attention maps) & 
Low (resolves mathematical\newline interpolation extremely fast) \\
\hline
{\bf Adaptation Role} &
Core pre-trained model per-\newline forming text generation &
Enhances task-specific fine-\newline tuning (e.g., via GenLoRA) by\newline synthesizing parameters\\
 \hline
\end{tabular}
\caption{Comparing standard LLM transformers with RBF network}
\label{table:kbt54xpq}
\end{table}

\section{Fast, high-accuracy RBF network without training} 

My model is a standard exact RBF interpolator. I call it \textcolor{index}{interpolator}\index{interpolator} rather than network as it does not involve DNNs. The 
 machinery is pretty heavy similarly to DNNs, with many shared features. However, it uses 10,000 fewer embeddings  
because I focus on corporate corpuses. That is,  SLMs trained on corpus and English language (a tiny fraction of the whole Internet)  to answer specialized business questions, instead of generic LLMs
that can write code, solve math problems, and answer any question in any language. 
The focus is on concise yet exhaustive and structured answers, with a relevancy score attached to each item in the response.  

For numerical data, I use radial \textcolor{index}{Gaussian mixtures} as seen in formulas~(\ref{totor}) and~(\ref{porc}). 
See also formula~(\ref{porcw}) for adaptation to text data. In the Python code, text strings are kept ``as is", and not even turned
into numerical \textcolor{index}{embeddings}, with \textcolor{index}{vector databases} replaced by nested hashes. Also, the choice of the kernel $K$ is not important. Instead, I put emphasis on fast computations using
pre-tabulated values with the minimum precision needed via \textcolor{index}{quantization}.   

\noindent The two main differences with standard models are as follows:
\vspace{1ex}
\begin{itemize}
\item The weights $w_k(x)$ depend on $x$ and are normalized: they add up to 1. This contrasts with most other implementations that do not require
 normalization and instead use DNNs to find the best weights (called parameters).
\item I focus on the singular case when $\tau\rightarrow\infty$ in~(\ref{totor}). 
It results in exact predictions on the training data,~no matter the weights and other parameters or hyperparameters. 
This is a consequence of theorem~\ref{thcarb} proved in chapter 6 in~\cite{vgnewai26}. To work well, it requires careful attention to the numerical analysis aspects.  
Then, no training is needed. 
\end{itemize}
\vspace{1ex}
\noindent There are several other differentiatiors. I use multi-tokens instead of embeddings, each consisting of a sequence of stemmed words. This allows to better match business acronyms in the prompt to text in the corpus.
Also, there are different types of multi-tokens: regular and contextual. The latter consists for instance of text elements found in titles, tags, categories, or bigger fonts in PDF documents. The user can specify tags and negative keywords when prompting. 

\subsection{Model description and formulation}\label{descr1}

Predictive models are typically denoted as $Y = f(X)$ where the response $Y$ is a column vector with $n$ observations, and the input data
$X$ is a table with $n$ rows and $m$ columns. The columns are called the \textcolor{index}{features}\index{feature (machine learning)}, and $m$ is the \textcolor{index}{dimension}\index{dimension}. Here I use the notation $\beta$ to represent $X$, with $\beta_k$ corresponding to the $k$-th row and $\beta_{kj}$ being the value in cell $(k, j)$ in the table.  The $\beta_k$'s are called the \textcolor{index}{nodes}\index{node (interpolation)}. More specifically, $\beta = \varphi(X)$ where $\varphi$ is an invertible transform (or a chain of invertible transforms) used as \textcolor{index}{normalizer}\index{normalization} to dramatically
 improve the performance. The model is as follows:

\begin{equation}
f_\text{pred}(x) = \sum_{k=1}^n \omega_k(x) \, f(\beta_k) \, \exp\Big[-\tau\, K(x, \beta_k)\Big],\label{totor}
\end{equation}
where $f_\text{pred}(x)$ is the predicted value of $f(x)$ for a function $f$ known only at 
$n$ locations $\beta_1,\dots,\beta_n$ in a space of dimension $m$.  
For all $x$, the \textcolor{index}{weights}\index{weights} $\omega_k(x)$ must satisfy: 
\begin{equation}
 \sum\limits_{k=1}^n \omega_k(x) \exp\Big[-\tau\, K(x, \beta_k)\Big] = 1.\label{wwc}
\end{equation}
\noindent Thus the weights depend on $x$ and implicitly on the nodes in a highly non-linear way. This is our first strong departure from many kernel models. 
Another major difference is that the number of non-zero weights can be as large as $n$, compared to other methods where most weights
 $\omega_k(x)$ are zero unless $x$ and $\beta_k$ are close enough. The function $K$ is called the \textcolor{index}{kernel}\index{kernel method}. 
It must be positive, symmetric, and equal to zero only if both arguments are identical. A typical example in our framework is
\begin{equation}
K(x, \beta_k) = \Big[\frac{1}{m} (x-\beta_k)(x-\beta_k)^T\Big]^\gamma = \Bigg[\frac{1}{m}\, \big|\big|x-\beta_k\big|\big|^2 \Bigg]^\gamma \label{porc}
\end{equation}
assuming $x, \beta_k$ are row vectors with $m$ components ($m$ as large as 1000), and $T$ is the transposition operator. So, the vector product in~(\ref{porc}) 
 is a \textcolor{index}{dot product}\index{dot product}. I mostly used $\gamma = \frac{1}{2}$, with $\gamma=1$ on occasions. 
The multiplication by $1/m$ in~(\ref{porc}) proves particularly useful when $m$ is large, acting as a normalizer. A slight generalizations consists of using
\begin{equation}
K(x, \beta_k) = \sum_{j=1}^m \theta_j \delta(x_j, \beta_{kj}) \label{porcw}
\end{equation}
where the $\theta_j$'s are positive and add up to 1. In case of numerical values, $\delta(x_j, \beta_{kj}) = (x_j - \beta_{kj})^2$.
In the context of LLMs, the meaning is as follows:
\vspace{1ex}
\begin{itemize}
\item Both $x$ and $\beta$ are text; $x$ comes from a prompt, while $\beta$ comes from a corpus.
\item $x_j, \beta_{kj}$ may be small text strings, and $\delta(x_j, \beta_{kj})$ is some association measure between $x_j$ and 
 $\beta_{kj}$, such as the \textcolor{index}{enhanced PMI}\index{PMI!enhanced PMI} in the xLLM model. 
\item $f(x) = P(x_m \,|\, x_1,\dots,x_{m-1})$ is the probability to observe $x_m$ at the end of a text string $x$, given that the previous text elements in the string are $x_{m-1}, x_{m-2},\dots, x_1$ in that order. Thus, it deals with next token prediction. 
\item The parameters $\theta_j$ weight each element of $x$ based on its position $j$ in the string $x$. Typically, the $\theta_j$'s are decaying weights to give more importance to word elements with a location closer to $x_m$ in the string $x$. 
\end{itemize}
\vspace{1ex}
Model~(\ref{porc}) is a particular case of~(\ref{porcw}) with all
$\theta_j$'s being equal. Ignoring the $\varphi$-transform for now, $\beta$ is called the \textcolor{index}{training set}, whether actual training is needed or not. The first important result, applying both to text and numerical data, is as follows:
\vspace{1ex}

\begin{theorem}\label{thcarb} If $x$ is an observation in the training set and $f(x)\neq 0$, then formula~(\ref{totor}) estimates $f(x)$ exactly when $\tau\rightarrow\infty$, whether
 based on kernel~(\ref{porc}) or~(\ref{porcw}), both for text or numerical data. No training is needed. 
\end{theorem}
\vspace{1ex}

\noindent Note that even if multiple nodes $\beta_k$ are duplicate but have the same value $f(\beta_k)$, the theorem remains true. Thus, the function $f$ can approximate any data, even the most chaotic ones, even if the data is a white noise. The convergence is pointwise, not uniform. Thus to make correct predictions outside the training set, you may need a large training set if $f$ is not smooth.

\subsection{Benign overfitting, other features and benefits}

\begin{figure}[H]
\centering
\includegraphics[width=10cm, height = 8cm]{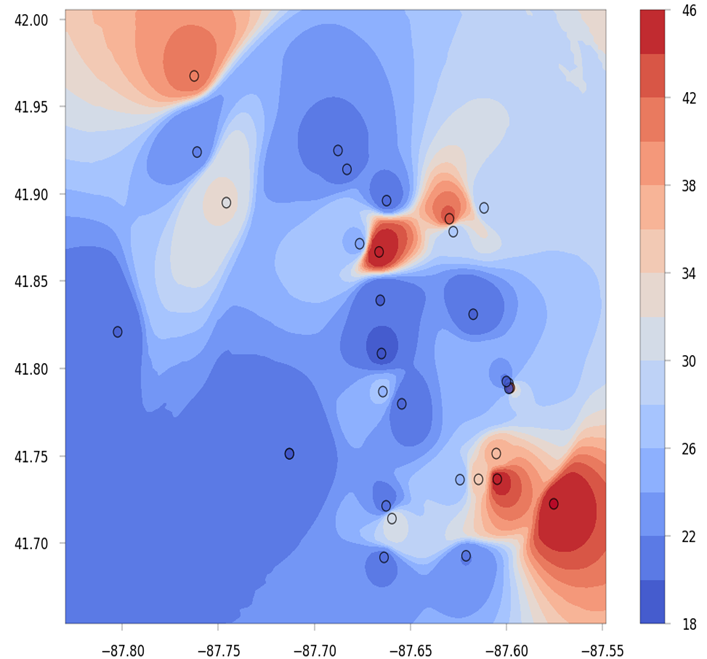}
\caption{Chicago temperatures with exact\\predictions on training set points (the dots)}
\label{fig:g6vgeo22}
\end{figure}

My model, jut like most DNNs, benefit from \textcolor{index}{benign overfitting}: the ability to get very good predictions outside the training set despite being 100\% correct (thus very overfit) on the training data. For DNNs, why it works remains a mystery. However, in my case, the explanation is straightforward. The model originates from exact spatial interpolation in 2 dimensions, also known
 as \textcolor{index}{kriging}. If you look at figure~\ref{fig:g6vgeo22}, the dots represent training set locations where
 the temperature is perfectly matched. Outside those locations, you still have good predictions, actually better than from standard models that tend to oversmooth, including irregular contour lines instead of smooth elliptic curves, to better represent local variations. The bottom right corner is the city of Chicago with its heat dome.

\noindent My model generalizes the 2D case to 10,000+ dimensions, without suffering from
the 
\textcolor{index}{curse of dimensionality}. The reason is because even though the training set occupies a tiny portion of the
space in high dimensions, this is also true for the validation set, and both domains significantly overlap. Because both deal
 with the same narrow type of data: what is in your specialized corpus for the most part. 
In addition, it is easy to understand why the model works, since it consists of locations and scale parameters, thus 
based on \textcolor{index}{explainable AI} instead of Blackbox parameters, with connection to 
\textcolor{index}{nearest neighbor} techniques. Corpuses where the same terms have different meanings depending on the business department, benefit from splitting into sub-LLMs, also known as \textcolor{index}{mixtures of experts}.

\noindent Other features of my model:
\vspace{1ex}
\begin{itemize}
\item {\bf Deterministic AI}. Standard LLMs rely on DNNs and probabilistic models, for instance Gaussian latent variables, Markov chains, autoregressive models, and
 stochastic gradient descent.  None of this is present in my model, leading to \textcolor{index}{replicability}. 
\vspace{1ex}
\item {\bf Temperature fine-tuning}. To enrich the response to a prompt, LLMs allow you to choose a temperature value to generate different answers when you run the same prompt twice. This feature is also available in my model, yet with full
 replicability as long as you choose the same seed.  Indeed, I use my own \textcolor{index}{NPG random generator}, faster and better, see \href{https://mltblog.com/npg}{here}. It gives you full control 
and replicability, unlike Blackbox PRNGs which may change over time without your knowledge. This happened recently in Python, with the Mersenne Twister transparently replaced by PCG64 and leading to different results even if using the same seed. 
\vspace{1ex}
\item {\bf Attention mechanism}. As in all transformer-based LLMs, my model incorporates an attention mechanism: the order of the words in a sentence does matter. See formula~(\ref{porcw}) and the associated description. 
\vspace{-1.5ex}
\item {\bf Three-way training}. Predictions are always correct on the training set, thus you cannot use the standard two-way 
 method (training + validation sets) to optimize my model. Instead, you need an intermediate set between training and validation. I call it the optimization set. You use it to fine-tune the hyperparameters. 
\vspace{-1.5ex}
\item {\bf Adaptive parameters}. The hyperparameters $\tau$ and $\gamma$ in formulas~(\ref{porc}) and~(\ref{porcw})
 can depend on $k$ instead of being static. In this enhanced model (the adaptive version), they play the role of parameters in standard DNNs. They are optimized using a DNN 
 on the optimization set described above, as the predictions are still 100\% on the training set. 
A good analogy are thermonuclear bombs (DNNs in this context) that need a standard nuclear detonation (my RBF device) 
as a pre-processing step to trigger the massive explosion.  
Alternatively, for the second step, you can use another RBF on top of the first one, instead of DNNs.  
\vspace{1ex}
\item {\bf Weights distillation}. The number $n$ of training set embeddings in formula~(\ref{porcw}) can be very large.
Yet most terms in the sum contribute very little. You can speed up computations by
dropping terms (or~$\beta_k$'s) where $x$ and $\beta_k$ are almost never close neighbors regardless of the input $x$ coming from the prompt,~by analyzing data over time to optimize for speed via self-learning. I call it \textcolor{index}{auto-distillation}. 
Another idea is to do searches in dimensions much lower than $m$ to reduce the number of terms on the flight, depending on $x$.
By contrast, LLM products on the market are very expensive as tons of negligible weights that could be ignored, cost you the same
 as high-value weights.    
\end{itemize}
\vspace{1ex}

\noindent Even though $n$ is large, typically in the high six or low seven digits, it is still far smaller than in standard LLMs where the 
lowest value that works is in the billions, due to the nature of DNNs. In short, the gain in efficiency is of the order 10,000. 
The main reason is because we are dealing with SLMs instead of LLMs. Also, I use long tokens consisting of stemmed words augmented with stemmed acronyms, further lowering $n$ while increasing relevancy and exhaustivity in the response, with a focus on concise structured output as opposed to long paragraphs and high-quality grammar. The end goal is summarization, not essay writing.  

\subsection{From billions to fewer than a million parameters}
As a side note, the number $V_N$ of unique tokens found in a corpus follows the \textcolor{index}{Heap law}: 
$V_N = \lambda N^\nu$, where $N$ is the total number of tokens, a linear function of the corpus size. 
In practice, $0.4< \nu < 0.6$ and $10<\lambda< 100$. 
If instead of using the whole Internet to train your model, you carefully select sources covering 0.01\%,
 you reduce $V_N$ by a factor 100 without impact on quality.  
Also, the vocabulary of an average human consists of less than 30k tokens. Thus using a lot more is pointless,
 as most will almost never be shown in an answer, and when they do, they won't be understood.

\begin{table}[H]
\centering
\renewcommand{\arraystretch}{1.0}
\begin{tabular}{cccc} 
 \hline
  &  &  &        \\  [-2.5ex]
  1--grams & 2--grams & 3--grams & 4--grams \\
 \hline 
\hline 
   &  &  &        \\  [-2ex]
5804 & $\num{46587}$ & $\num{66625}$ & $\num{74717}$ \\
 \hline
\end{tabular}
\caption{$m$-gram counts, NVIDIA corpus}
\label{table:kb8h3rbabare2e}
\end{table}

\noindent The same probabilistic laws apply to unique token combinations consisting of $m$ tokens (embeddings, in short). 
While the number quickly explodes as a function of $m$ when $m$ is small, it also quickly tappers off  
when $m$ reaches some rather low critical threshold, depending on corpus size. This is illustrated in table~\ref{table:kb8h3rbabare2e}, where $m$-gram
stands for multi-token with $m$ words. 

The reason is as follows: the immense
majority of multi-tokens of (say) size $m=10$ is absent in the corpus, and the few that show up have very low
 frequency (typically less than 2 occurrences) making them useless for predictions, and best ignored. 
There will definitely be very long multi-token ($m$ large) that occur a few times, and it's best to keep them. 
All this to explain why my model needs much fewer tokens, and why it makes sense to use hashes rather than vectors to store them, 
to efficiently handle \textcolor{index}{variable-length embeddings} and sparsity. 


\section{Case study: 96\% correct prediction rate}

In one tough test involving numerical \textcolor{index}{synthetic data}, I scrambled a training set with $n=10^4$ observations~and $m=10^3$ dimensions
 by adding significant noise. I then dropped 70\% of the observations in the training set and computed
$f_\text{pred}$
 with~(\ref{totor}).
The R-squared on the full training data with blurry response dropped from 100\% to 50\%, mostly due to the noise.
Yet, outside the training set, on input data with known noise-free response, it was 97\%. 
Note that since the data is synthetic, $f_\text{real}$ is known.  
In this case, $f_\text{real}$ consisted of 100 irregular splines---far from a smooth function---and I used $\tau=500$ as an
approximation to the desired $\tau=\infty$. In short, my system was able to recover the original signal. It also shows
 that in specific situations, it is possible to get a good outcome out of bad data. Even in high dimensions. 

This is one of many examples that seem to defeat all the known laws of statistics, used in my \textcolor{index}{stress-test} and summarized in table 6.1 in~\cite{vgnewai26}. The fully replicable Python code is also available in that book. It involves normalization and recalibration techniques, 
central to DNNs. The case I now discuss deals with text rather than numerical data. It is less challenging despite
 coming from a real-word corpus (NVIDIA) instead of AI-generated data.

\subsection{NVIDIA case study}

 
The mechanics for text data processing are described in detail in section 6.2.3 in~\cite{vgnewai26}. Here I provide a quick overview and the main results. The ability to provide different answers to a same prompt comes from the fact
 that $f_\text{pred}$ returns more than just the predicted value, but also other multi-tokens $\beta_k$ that significantly contribute
 to the sum in~(\ref{totor}). Also, the correct prediction rate is significantly boosted outside the training set by a 
mechanism to avoid $f(x)=0$ and the corresponding issue in theorem~\ref{thcarb}.

\begin{figure}[H]
\begin{minipage}{.50\textwidth}
\centering
\captionsetup{justification=centering}
\vspace{0.25ex}
\includegraphics[height=6cm]{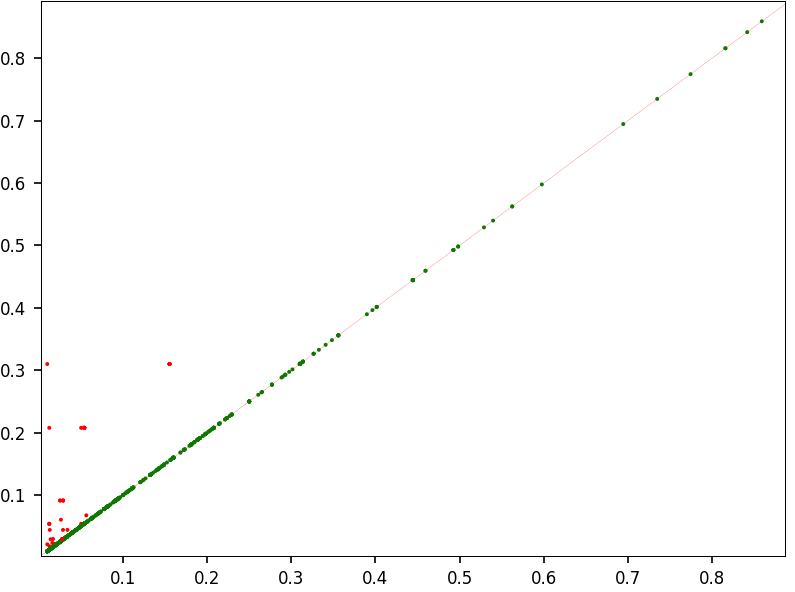}
\vspace{-1ex}
\caption{{\small Goodness of fit when predicting the next token, with correct prediction (green dots) on red diagonal}}
\label{fig:aajj5r4ewkigfi7}
\end{minipage}
\begin{minipage}{.50\textwidth}
\centering
\captionsetup{justification=centering}
\includegraphics[height=6cm]{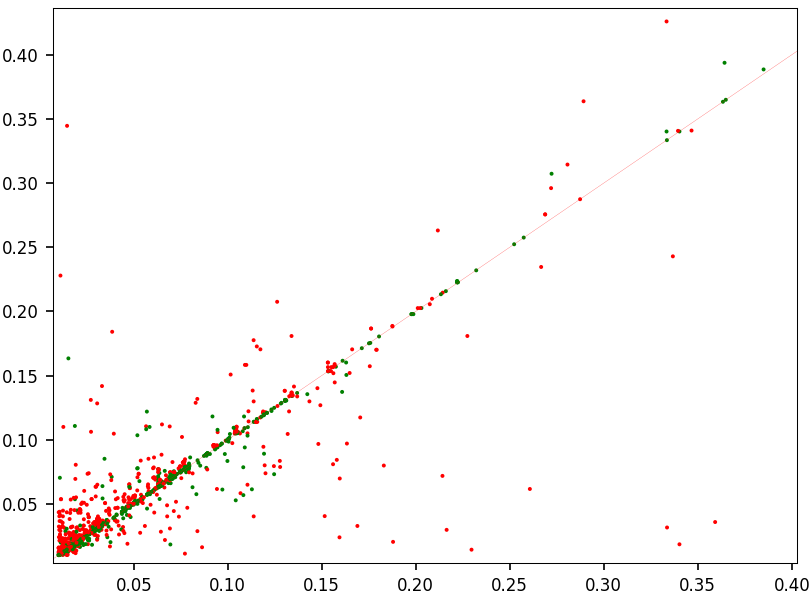}
\caption{{\small Goodness of fit when looking for related multi-tokens, perfect fit for green dots on red diagonal}}
\label{figggfrtfcfl7bd3csku6}
\end{minipage}
\end{figure}

\noindent I tested my model for two different tasks: predicting the next token, and finding relevant multi-tokens associated to
 the one you are interested in (denoted as $x$). The latter is fundamental to suggest related prompts to a user querying the system. 
Performance outside the training set is shown respectively in figures~\ref{fig:aajj5r4ewkigfi7} and~\ref{figggfrtfcfl7bd3csku6}. 
Again, I use the terms multi-token and embeddings interchangeably. However, my ``single'' tokens consist of whole stemmed words instead of classic, small tokens. 

\begin{figure}[H]
\begin{minipage}{.50\textwidth}
\centering
\captionsetup{justification=centering}
\includegraphics[height=6cm]{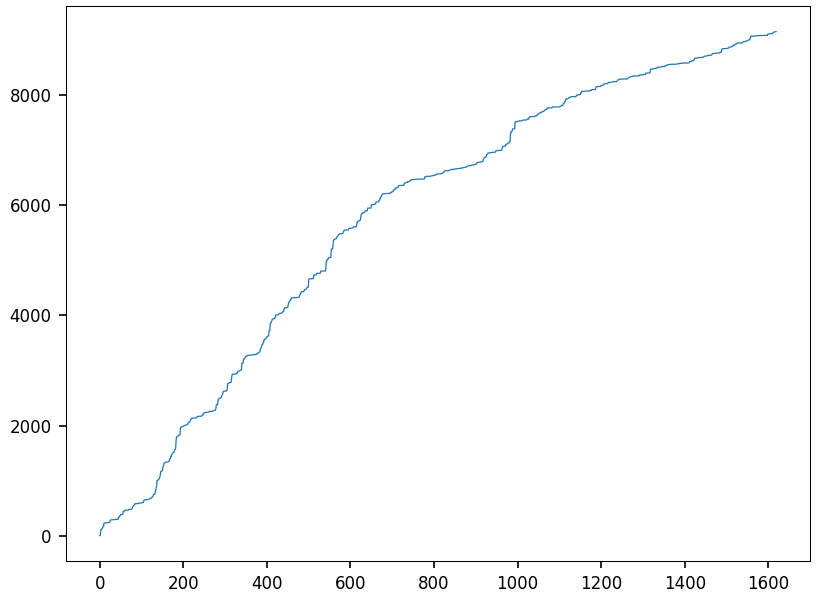}
\caption{{\small Cumulative number of nodes triggered  (Y-axis) out of 15,000,  based on cumulative queries on the X-axis}}
\label{fig:aajj5r4ewkigfi7fdsx2}
\end{minipage}
\begin{minipage}{.50\textwidth}
\centering
\captionsetup{justification=centering}
\includegraphics[height=6cm]{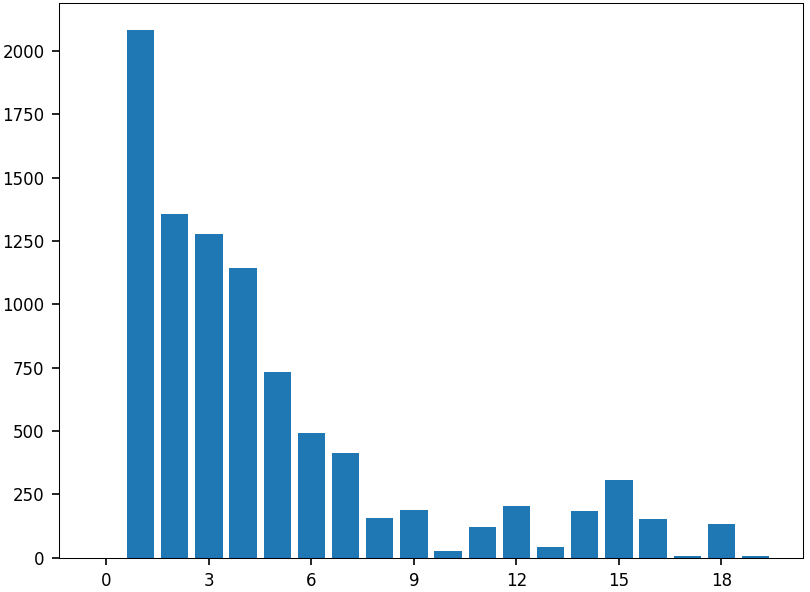}
\caption{{\small Nodes  triggered  out of 15,000 to answer a\\ query (X-axis), distribution based on 9,000 queries}}
\label{figggfrtfcfl7bd3csku654z10}
\end{minipage}
\end{figure}

Here $m=4$. The number of multi-tokens of size $m$ in the training set is $n=15,000$, out of the 74,717~total for the corpus as reported in 
table~\ref{table:kb8h3rbabare2e}. To predict the next token based on the 3 previous ones, 
 I test all the 5,804 single tokens in the corpus to find which one maximizes $f_\text{pred}$. 
I also run a number of queries to see how many unique multi-tokens $\beta_k$ from the training set get triggered
 in~(\ref{totor}) to answer them. See figure~\ref{fig:aajj5r4ewkigfi7fdsx2} with the number of queries on
the X-axis. To serve 1,600 queries, you need about 9,000 $\beta_k$'s, that is, about 60\% of the training set size. 
That number initially increases linearly with the number of queries, but then taper off.  

An even stronger indicator of the sparsity of the system is shown in figure~\ref{figggfrtfcfl7bd3csku654z10}:
 each query $x$ in a sample of 9,000 triggers fewer than 20 $\beta_k$'s out of 15,000 terms in~(\ref{totor}). 
Of course which ones are triggered depend on $x$. But it shows that there is room 
for improvement to potentially significantly reduce the number of terms in~(\ref{totor}) by 
only keeping the active ones, either  depending on the prompt (big saving $>99.5\%$ but difficult), or globally (easy but smaller saving $<40\%$).

\noindent Table~\ref{table:kb8h3rdv02e} shows examples of related multi-tokens associated to an
 original multi-token $x$ with $m=4$ stemmed~words, along with metrics ($f(x)$ and grade) assessing the quality
of the match, based on the corpus.  
Table~\ref{table:kb8g4zpoe} shows the prediction $\text{Next}_\text{pred}$ based on the 3 previous tokens, 
and compare it with the real one $\text{Next}_\text{real}$ that was deleted from the multi-token in question.  
I focused on the few cases outside the training set where the prediction is not correct.  
The suggested next token, while not correct, is actually very relevant, enriching the response. 
\vspace{1ex}

\begin{table}[H]
\centering
\renewcommand{\arraystretch}{1.0}
\small
\begin{tabular}{llcc} 
 \hline
  &  &  &        \\  [-2.5ex]
  Original multi-token & Related multi-token & $f(x)$ & Grade \\
 \hline 
\hline 
   &  &  &        \\  [-2ex]
california$\sim$februari$\sim$24$\sim$2023 & sec$\sim$februari$\sim$24$\sim$2023 & 0.0339 & A \\
larg$\sim$languag$\sim$model$\sim$cloud & larg$\sim$languag$\sim$model$\sim$nvidia & 0.0204 & A\\
larg$\sim$languag$\sim$model$\sim$cloud & larg$\sim$languag$\sim$model$\sim$perform & 0.0204 & A\\
larg$\sim$languag$\sim$model$\sim$cloud & custom$\sim$languag$\sim$model$\sim$gener & 0.0112 & B\\
compens$\sim$data$\sim$center$\sim$infrastructur &  cost$\sim$data$\sim$center$\sim$infrastructur & 0.1052 & A\\
may$\sim$neg$\sim$impact$\sim$financi & materi$\sim$advers$\sim$impact$\sim$financi & 0.0132 & B\\
cockpit$\sim$electr$\sim$vehicl$\sim$comput & solut$\sim$electr$\sim$vehicl$\sim$maker & 0.0152 & B\\
fiscal$\sim$2023$\sim$2022$\sim$audit & year$\sim$2023$\sim$2022$\sim$2021 & 0.0129 & B\\
decis$\sim$execut$\sim$compens$\sim$fiscal & highlight$\sim$execut$\sim$compens$\sim$program & 0.0157 & B\\
purpos$\sim$pay$\sim$ratio$\sim$calcul & employe$\sim$pay$\sim$ratio$\sim$wa & 0.0100 & B\\
arrang$\sim$match$\sim$contribut$\sim$401 & huang$\sim$match$\sim$contribut$\sim$401 & 0.0286 & A\\
arrang$\sim$match$\sim$contribut$\sim$401 & 2022$\sim$match$\sim$contribut$\sim$401 & 0.0286 & A\\
goal$\sim$fiscal$\sim$2023$\sim$revenu & result$\sim$fiscal$\sim$2023$\sim$revenu & 0.0170 & A\\
 \hline
\end{tabular}
\caption{Sample multi-tokens with their top relatives and quality metrics}
\label{table:kb8h3rdv02e}
\end{table}


\begin{table}[H]
\centering
\renewcommand{\arraystretch}{1.0}
\small
\begin{tabular}{lllcc} 
 \hline
  &  &  &    &    \\  [-2.5ex]
  First 3 tokens & Next$_\text{real}$ & Next$_\text{pred}$ & $f(x)$ & Grade \\
 \hline 
\hline 
   &  &  &    &    \\  [-2ex]
paid$\sim$neo$\sim$fiscal & 2021 & 2022 & 0.0540 & 0.01247\\
incom$\sim$abov$\sim$fiscal & 2021 & year &  0.2078 & 0.01247\\
rule$\sim$make$\sim$proxi & materi & statement & 0.0443 & 0.02880\\
total$\sim$defer$\sim$tax & asset & benefit & 0.0218 & 0.01813\\
unreal$\sim$loss$\sim$januari & 29 & 30 & 0.3100 & 0.15540\\
decreas$\sim$fiscal$\sim$year & 2022 & end & 0.0298 & 0.01693\\
subject$\sim$outstand$\sim$stock & option & award & 0.0914 & 0.02889\\
employe$\sim$nvidia$\sim$corpor & subsidiari & govern & 0.0677 & 0.05647\\
 \hline
\end{tabular}
\caption{Examples of failed prediction for the next token}
\label{table:kb8g4zpoe}
\end{table}


\subsection{Next token prediction: computational complexity}

The computational complexity of predicting the next token using a Deep Neural Network (DNN)—typically an autoregressive transformer—is \(\mathcal{O}(L \times d^2)\) for each token, where \(L\) is the number of layers and \(d\) is the hidden dimension. Total inference cost scales quadratically with sequence length due to attention mechanisms.

The computational complexity after the training phase—meaning during inference (running the model to predict the next token)—is \(\mathcal{O}(L \cdot d \cdot (T^2 + V))\) for a standard transformer model. When utilizing optimization techniques like key-value (KV) caching, the per-token computational complexity 
reduces to \(\mathcal{O}(L \cdot d^2 + L \cdot T \cdot d)\). I now breakdown the post-training computations
 with typical values for small models (SLMs) to assess actual performance. The numbers come from Google AI. 
\vspace{1ex}
\begin{itemize}
\item $V$ is the vocabulary size, that is, the number of unique tokens. In my small case study with stemmed tokens, $V=5804$, see table~\ref{table:kb8h3rbabare2e}. Typical values range from 32,000 and 50,000. 
\item $T$ is the context length and range from 4,096 to 8,192 tokens. In my case study, $T=m=4$. My model~can handle far larger values (when it makes sense) and has been tested on numerical data with $m=1000$.
\item $L$ is the number of layers and range from 24 to 32.
\item $d$ is the hidden dimension and range from  2048 to 3073. 
\end{itemize}  
\vspace{1ex}
\noindent The value for $T$ is very small in my model compared to standard transformer-based SLMs because the goal is different: identifying related concepts, as opposed to producing full, long English sentences. That aside, the computational complexity of my model, 
without optimization, is  $\mathcal{O}( n \cdot T \cdot V)$, where $n=15,000$ is the number of unique multi-tokens of size $T=4$ in the training set.   
For the tasks performed here, my system is as efficient as standard DNNs while entirely skipping the costly training phase. Optimization techniques consist in
 finding how to skip most of the irrelevant terms (out of $n$) when computing~(\ref{totor}) for a specific $x$, and using pre-tabulated values
 in a mechanism similar to KV cache. 

\subsection{Earlier DNN-free model with exact predictions on training set}

The first model tested for numerical data involved predicting $f(x)$ by first approximating $x$ with a linear convex combination of the nodes
$\beta_1,\dots,\beta_n$ in the training set, solving the quadratic constrained optimization problem   
\begin{equation}
\omega^*(x) = \arg\min_\omega \, \Big|\Big|x - \sum\limits_{k=1}^n \omega_k \beta_k\Big|\Big|^2 \label{bobres}
\end{equation}
where $\omega = (\omega_1, \dots, \omega_n)$ and the positive weights $\omega_k$ add up to 1. 
If $x=\beta_i$ is one of the nodes, then $\omega^*_k(x)$ is zero for all $k$ except $k=i$ for which $\omega^*_k(x) = 1$.
Then, instead of~(\ref{totor}), I predict $f(x)$ with
\begin{equation}
f_\text{pred}(x) = \sum\limits_{k=1}^n \omega^*_k(x) f(\beta_k)
\end{equation}
Again, it leads to exact predictions $f_\text{pred}(x)=f(x)$ in any dimension when $x$ is in the training set, combined with the desirable
 feature known as \textcolor{index}{benign overfitting}. However, the optimum is reached when all the weights $\omega^*_k(x)$ are zero except for the nodes $k$ where $\beta_k$ is one of the $m+1$ vertices in the polyhedron encompassing $x$ in the $m$-dimensional 
 \textcolor{index}{Delauney triangulation}\index{Delauney triangulation} of
 the training set. I did not pursue this approach since I did not see how to adapt it to LLMs. 
The other drawback is computational complexity, as the optimization problem must be solved for each new vector $x$.
In LLMs, $x$ would be some embedding coming from a prompt.


\section{Conclusions}

My alternative to DNNs for LLM architecture may have been perceived as an isolated, 
one-off model untested by others 12 months ago. 
With Chinese researchers now actively working on the exact same model, it is becoming 
 a topic of significant interest. They call it ``RBF networks" while I used the word ``kernel method" in the past. Both terms are correct 
and widely known in contexts other than LLMs. The difference reflects the research field you are coming from,  
 but both points to the exact same equations. However, my approach is unique in the sense that it does not use DNNs 
 to compute the weights. Instead, I obtain them in one-shot without training,
with 100\% correct prediction on the training set, without bad overfitting, in high dimensions. 

I introduce auto-distillation and pre-tabulated values as mechanisms to speed up computations. I
also discuss why it works with 10,000 fewer embeddings. In the original book where my method was first
published~\cite{vgnewai26}, I also discuss distillation-resistant invisible watermarking techniques to protect your model. 
Last but not least, I feature a case study with 96\% correct prediction rate for next token, and
 discuss replicability, explainability and deterministic AI attached to the model, with the ability to allow
 for controlled randomness in the response if desired. Due to perfect predictions on the training set,
 I explain how to perform three-way training to fine-tune the hyperparameters. The 96\% correct prediction rate outside the training set is far above the 30 to 55\% achieved by standard transformer-based models,
 while avoiding costly training and without increased compute time post-training.  
This high performance is due to specialization to the specific corpus, by contrast to generic predictors.

The next steps include working with a larger corpus, and performing tasks beyond predicting the next token, suggesting relating queries, or finding synonyms. The methodology is also well suited for image classification and problems with numerical data (time series and so on).


\bibliographystyle{plain} 
\bibliography{refstats} 

\hypersetup{linkcolor=red} 
\hypersetup{linkcolor=red}

\end{document}